\documentclass[letterpaper, 10 pt, conference]{ieeeconf}  
\usepackage[left=0.75in,right=0.75in,top=0.75in,bottom=0.75in]{geometry}

\IEEEoverridecommandlockouts                              

\usepackage[dvipdfmx]{graphicx}
\usepackage{amsmath}
\usepackage{comment}
\usepackage{url}
\usepackage{color}
\usepackage{subfigure}
\usepackage{amsmath}
\usepackage{mathtools}
\usepackage{cite}
\usepackage{algorithm}
\usepackage{algorithmic}
\usepackage{multirow}

\title{\LARGE \bf
Efficient Solution to 3D-LiDAR-based Monte Carlo Localization with Fusion of Measurement Model Optimization via Importance Sampling
}

\author{Naoki Akai$^{1,2}$
\thanks{$^{1}$Naoki Akai is with the Graduate School of Engineering, Nagoya University, Nagoya 464-8603, Japan {\tt\small akai@nagoya-u.jp}}%
\thanks{$^{2}$Naoki Akai is with the LOCT Co., Ltd., Nagoya 464-0805}%
}

\newcommand{\argmax}{\mathop{\rm argmax}\limits}
\newcommand{\argmin}{\mathop{\rm argmin}\limits}

\begin{document}

\newcommand{\1}{\mbox{1}\hspace{-0.25em}\mbox{l}}
\renewcommand{\baselinestretch}{1.0}

\maketitle
\thispagestyle{empty}
\pagestyle{empty}

\begin{abstract}

This paper presents an efficient solution to 3D-LiDAR-based Monte Carlo localization (MCL). MCL robustly works if particles are exactly sampled around the ground truth. An inertial navigation system (INS) can be used for accurate sampling, but many particles are still needed to be used for solving the 3D localization problem even if INS is available. In particular, huge number of particles are necessary if INS is not available and it makes infeasible to perform 3D MCL in terms of the computational cost. Scan matching (SM), that is optimization-based localization, efficiently works even though INS is not available because SM can ignore movement constraints of a robot and/or device in its optimization process. However, SM sometimes determines an infeasible estimate against movement. We consider that MCL and SM have complemental advantages and disadvantages and propose a fusion method of MCL and SM. Because SM is considered as optimization of a measurement model in terms of the probabilistic modeling, we perform measurement model optimization as SM. The optimization result is then used to approximate the measurement model distribution and the approximated distribution is used to sample particles. The sampled particles are fused with MCL via importance sampling. As a result, the advantages of MCL and SM can be simultaneously utilized while mitigating their disadvantages. Experiments are conducted on the KITTI dataset and other two open datasets. Results show that the presented method can be run on a single CPU thread and accurately perform localization even if INS is not available.

\end{abstract}


\section{Introduction}
\label{sec:introduction}

Monte Carlo localization (MCL)~\cite{Thrun:2005:PR:1121596} is a well known approach to robust localization.
MCL robustly works if candidate poses, i.e., particles, are exactly sampled around the ground truth.
An inertial navigation system (INS) such as odometry and IMU can be used for accurate sampling, but many particles are still needed to be used for solving the 3D localization problem even if INS is available.
In particular, huge number of particles are necessary if INS is not available and it makes infeasible to perform 3D MCL in terms of the computational cost.
Scan matching (SM), that is optimization-based localization, efficiently works even if INS is not available because movement constraints of a robot and/or device is ignored in the optimization process.
However, SM sometimes determines an infeasible estimate against the movement.
We consider that MCL and SM have complemental advantages and disadvantages.
This paper presents a seamless fusion method of them that leverages advantages of MCL and SM while mitigating their disadvantages.
Here, the seamless fusion means that their are no heuristic methods for the fusion except for some approximation of probabilistic distributions.
In addition, the presented method enables to run 3D-LiDAR-based MCL on a single CPU thread even if INS is not used.

\begin{figure}[!t]
    \begin{center}
        \includegraphics[width = 85 mm]{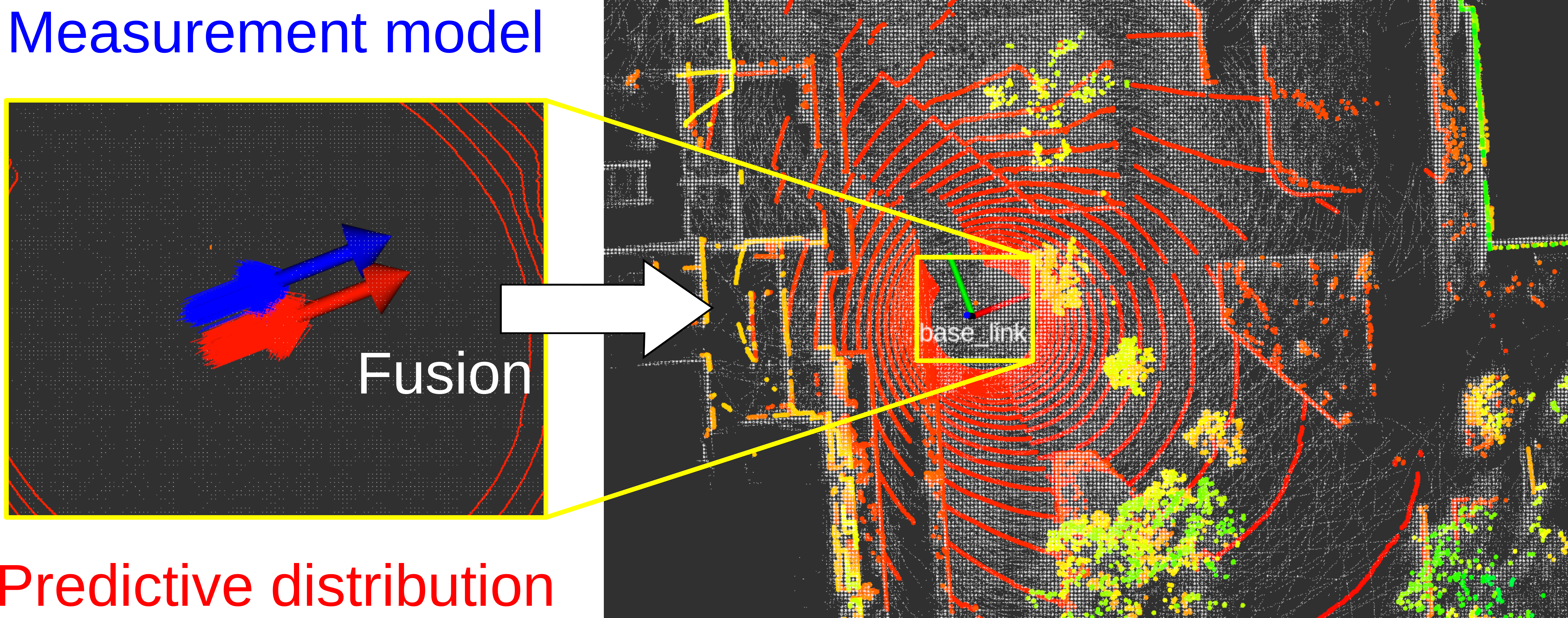}
        \caption{The presented method uses two probabilistic distributions to draw particles. The small red and blue arrows depict the particles drawn from the predictive distribution and measurement model. Predictive-distribution-based sampling is inaccurate if INS is not available, and measurement-model-based sampling is weak to measurement noises. However, these methods have different type advantages. The presented method leverages the advantages by fusing them with the measurement model optimization and importance sampling. The software used in this work is publicly available \protect\url{https://github.com/NaokiAkai/mcl3d_ros}.}
        \label{fig:concept}
    \end{center}
\end{figure}

To achieve the seamless fusion, we use importance sampling.
The use of importance sampling enables us to use multiple distributions to sample the particles.
In the proposal, we use predictive distribution and measurement model for the sampling.
In the typical particle filter (PF), particle poses are updated based on motion information.
The updated particles approximate a probabilistic distribution and it is referred to predictive distribution.
Hence, the updated particles can be regarded as sampled poses from the predictive distribution.
Sampling from the measurement model is a not trivial task.
To perform the sampling, we try to approximate the measurement model.
For the approximation, we use numerical optimization of the measurement model and this process is referred to SM in the proposal.
The problem for the measurement model optimization is that there are non differentiable factors owing to ad hoc implementation of measurement models.
For example in the likelihood field model (LFM), that is the popular measurement model~\cite{Thrun:2005:PR:1121596}, a distance field is used and it has a non differentiable term.
We will achieve measurement model optimization using numerical differential and it is effectively performed with the distance field representation method presented in~\cite{AkaiIV2020}.
Based on the optimization result, we approximate the measurement model as a normal distribution, where the Hessian matrix calculated in the optimization process is used to approximate the covariance.
Then, particles are sampled from the normal distribution and we achieve sampling from the measurement model.
Finally, the sampled particles are fused with the particles sampled from the predictive distribution based on the importance sampling manner.

Figure~\ref{fig:concept} shows benefits of the proposal.
PF-based localization is robust to noises and the presented method is also robust to miss estimate of SM.
In addition, SM cannot usually recover once optimization has failed since next estimate starts from an incorrect state.
However, the presented method mitigates influence of SM's miss estimate and enables to perform SM again.
Moreover, standard PF cannot work well to solve the 3D localization problem if INS is not available; however, the presented method works with small number of particles, e.g., 1000, owing to the fusion of SM.
The contribution of this paper is to propose a new framework to seamlessly fuse MCL and SM and provides an efficient solution for 3D-LiDAR-based MCL.
We believe that the new solution could make further contribution that is to extend 3D localization to other simultaneous state estimation problems such as we presented previously~\cite{AkaiJFR2023}.

The rest of this paper is organized as follow.
Section~\ref{sec:related_work} summarizes related works.
Section~\ref{sec:proposed_method} and \ref{sec:implementation} describe the proposed framework and its implementation way.
Section~\ref{sec:experiments} describes experimental conditions and results.
Section~\ref{sec:conclusion} concludes this study.

\section{Related work}
\label{sec:related_work}

3D-LiDAR-based localization is widely used for many robotic applications.
Iterative closet point (ICP) scan matching~\cite{Besl:1992:MRS:132013.132022} is a popular localization method.
However, ICP is often pointed out as inefficient and non robust to environment changes.
Normal distributions transform (NDT) scan matching~\cite{biber_iros2003:_ndt} is also a popular method and it is reported that NDT is efficient and robust more than ICP~\cite{Comparison_ICP_NDT}.
However, these methods are not considered as enough perfect for all the robotic applications.
Many authors have presented many types of localization methods.

Generalized iterative closest point (GICP)~\cite{GICP} is a generalized method of ICP and is considered as ICP that performs local-shape-to-local-shape matching.
GICP does not explicitly separate the point cloud unlike NDT.
Extension of GICP is presented by some authors, e.g.,~\cite{multi-channel-GICP}.
Honda {\it et al}. incorporate tiny neural networks into GICP and improve the data association metric and cost function using local geometric features~\cite{HondaRA-L2022}.
Our proposal does not perform local feature matching, but it performs measurement model optimization.
If a probabilistic model for local features are considered, it will be similar to our proposal.

LOAM~\cite{LOAM} is widely used in recent robotic applications.
In LOAM, LiDAR measurements are divided into local features such as edges and planes and these are used for scan registration.
LOAM achieves fast scan registration, but its accuracy depends on accurate point type labeling.
In addition, LOAM is not formulated to perform posterior estimation like PF does.
LOAM has several extensions such as LeGO-LOAM~\cite{legoloam2018}, but these are also implemented based on optimization.

Our proposal employs measurement model optimization that estimates a pose maximizing the measurement model using a distance field.
The similar method is presented by Caballero and Merino~\cite{DLL}.
In their method, the summation of distances obtained from the distance field is minimized.
This approach is similar to ICP; however, their method is more efficient and robust owing to the use of the distance field.
Our method performs similar optimization, but we maximize the measurement model using the Gauss-Newton method and this enables to approximate the measurement model with a normal distribution.
Owing to the approximation, we achieve seamless fusion of scan matching, i.e., the measurement model optimization, with MCL.
In addition, the authors used interpolation of the distance field to perform analytic gradient computation in~\cite{DLL}, but we use numerical gradient computation with the efficient distance field representation method~\cite{AkaiIV2020}.

Chen {\it et al}. presented learning-based observation model and range image-based MCL using 3D point clouds~\cite{chen2020iros, chen2021icra}, but they consider the 2D pose estimation problem.
In addition, the method presented in~\cite{chen2021icra} requires a graphics card to capable OpenGL.
In~\cite{Perez-GrauIJARS2017}, MCL with 3D LiDAR for aerial robots is presented.
The method uses a visual odometry and IMU and roll and pitch angles are directly obtained from IMU.
Hence, 3D position and yaw angle are estimated by matching the LiDAR measurements and they do not consider multiplication of the measurement models.
These increase computational efficacy.
A GPU accelerated localization method presented in~\cite{CUDAMCL}.
Indeed GPU is suitable for accelerating PF-based estimate; however, the presented method does not use GPU and achieves real time localization on a single CPU thread even though 3D pose is estimated by 3D LiDAR matching.

The similar problem that we tread in this paper is addressed in~\cite{https://doi.org/10.48550/arxiv.2302.06843}.
Adurthi proposed an integration method of PF and SM to solve the LiDAR-only localization problem.
We also solve the same problem and uses both PF and SM; however, a fusion method used in our proposal is different from that of described in~\cite{https://doi.org/10.48550/arxiv.2302.06843}.
The method presented in~\cite{https://doi.org/10.48550/arxiv.2302.06843} performs SM against only the particle that have the highest weight and its pose is only modified.
Since the pose modified particle usually has large likelihood, the method works even when INS is not available.
Our fusion method obeys more rigorous probabilistic manners owing to the use of the measurement model optimization and importance sampling and achieves more natural fusion that does not use heuristic methods except for some approximation of probabilistic distributions.

\section{Proposed method}
\label{sec:proposed_method}

\subsection{Problem setting}

Our objective is to estimate a 3D robot and/or device pose ${\bf x}$ using a sequence of 3D LiDAR measurements ${\bf z}_{1:t}$ and a given point cloud map ${\bf m}$.
This problem is formulated as a problem to estimate the posterior over the pose conditioned on the LiDAR measurements and map, and we focus on how to solve the problem using PF\footnote{The posterior over the pose can be estimated using other filters such as Kalman filter, but we would like to apply PF because our final objective is to extend the approach presented in~\cite{AkaiJFR2023} to the 3D localization problem.}.
A sequence of control input ${\bf u}_{1:t}$, such as odometry and IMU measurements, can also be used, but we assume that the use of the control input is optional.
If the control input is available, the pose is predicted based on the input, otherwise the pose is predicted based on previous estimates or is not predicted\footnote{If the pose is not predicted, the particles are needed to be expanded with other heuristic methods such as random sampling.}.
This difference yields formulation difference, but implementation difference is small.
In this section, we consider a case where the control input is not available.
If the control input is available, formulation is to be almost the same that of described in~\cite{Thrun:2005:PR:1121596}.
However, the proposal, i.e., fusion of SM via importance sampling, is not presented in~\cite{Thrun:2005:PR:1121596}.

\subsection{Formulation}
\label{subsec:formulation}

We aim to estimate the posterior over the pose denoted as $p({\bf x}_{t} | {\bf z}_{1:t}, {\bf m})$ using PF, but we first describe how the posterior is calculated.
The Bayes theorem is first applied using ${\bf z}_{t}$
\begin{align}
    p({\bf x}_{t} | {\bf z}_{1:t}, {\bf m}) = \eta p({\bf z}_{t} | {\bf x}_{t}, {\bf z}_{1:t-1}, {\bf m}) p({\bf x}_{t} | {\bf z}_{1:t-1}, {\bf m}),
    \label{eq:posterior_bayes}
\end{align}
where $\eta$ is a normalization constant.
$p({\bf z}_{t} | {\bf x}_{t}, {\bf z}_{1:t-1}, {\bf m})$ can be re-written as $p({\bf z}_{t} | {\bf x}_{t}, {\bf m})$ by applying D-separation~\cite{Bishop:2006:PRM:1162264} and it is referred to a measurement model~\cite{Thrun:2005:PR:1121596}.
Then, the law of total probability is applied to $p({\bf x}_{t} | {\bf z}_{1:t-1}, {\bf m})$ twice using ${\bf x}_{t-1}$ and ${\bf x}_{t-2}$ and Eq.~(\ref{eq:posterior_bayes}) is re-written as
\begin{align}
    \begin{gathered}
        p({\bf x}_{t} | {\bf z}_{1:t}, {\bf m}) = \eta p({\bf z}_{t} | {\bf x}_{t}, {\bf m}) \int \int p({\bf x}_{t} | {\bf x}_{t-1}, {\bf x}_{t-2}) \\
        \cdot p({\bf x}_{t-1} | {\bf x}_{t-2}, {\bf z}_{1:t-1}, {\bf m}) p({\bf x}_{t-2} | {\bf z}_{1:t-2}, {\bf m}) {\rm d}{\bf x}_{t-2} {\rm d}{\bf x}_{t-1},
    \end{gathered}
    \label{eq:recursive_bayes_filter}
\end{align}
where $p({\bf x}_{t} | {\bf x}_{t-1}, {\bf x}_{t-2})$ is a motion model using two previous poses.
This distribution can be approximated by particles that are predicted their poses based on previous two estimates such as linear interpolation.
The double integral term shown in Eq.~(\ref{eq:recursive_bayes_filter}) is referred to predictive distribution.

In our previous work~\cite{AkaiIROS2018}, we proposed the class conditional measurement model and showed that it is effective for robust localization against environment changes.
Thus, we replace the measurement model as
\begin{align}
    p({\bf z}_{t} | {\bf x}_{t}, {\bf m}) = \sum_{{\bf c}_{t} \in \mathcal{C}} p({\bf z}_{t} | {\bf x}_{t}, {\bf c}_{t} {\bf m}) p({\bf c}_{t}),
    \label{eq:class_conditional_measurement_model}
\end{align}
where ${\bf c}_{t}$ is a measurement class assigned to each measurement and $\mathcal{C}$ is a set of measurement labels.
In this work, two measurement labels $\mathcal{C} \in \{ {\rm known}, {\rm unknown} \}$ that indicate whether a measured obstacle exists on a map or not are used.

In the proposal, measurement model optimization is performed to find a pose that maximizes Eq.~(\ref{eq:class_conditional_measurement_model}).
However, we will discuss how $p({\bf z}_{t} | {\bf x}_{t}, {\bf m})$ is optimized in this section for generality.
In the following discussion, $p({\bf z}_{t} | {\bf x}_{t}, {\bf m})$ can be replaced to $\sum_{{\bf c}_{t} \in \mathcal{C}} p({\bf z}_{t} | {\bf x}_{t}, {\bf c}_{t} {\bf m}) p({\bf c}_{t})$.

\subsection{Measurement model optimization} 
\label{subsec:measurement_model_optimization}

Measurement-model-optimization-based localization estimates the pose that maximizes the measurement model.
The measurement model, $p({\bf z}_{t} | {\bf x}_{t}, {\bf m})$, is decomposed by assuming that each sensor measurement is independent one another.
The pose maximizing the measurement model is denoted as
\begin{align}
    {\bf x}_{t} = \argmax_{{\bf x}_{t}} \prod_{k=1}^{K} p({\bf z}_{t}^{[k]} | {\bf x}_{t}, {\bf m}),
    \label{eq:optimization_prod}
\end{align}
where $K$ is the number of measurements and ${\bf z}_{t}^{[k]}$ is $k$th measurement in ${\bf z}_{t}$.
Since many times of multiplication yields an unstable computation result, we re-write Eq.~(\ref{eq:optimization_prod}) using the negative logarithm
\begin{align}
    {\bf x}_{t} = \argmin_{{\bf x}_{t}} \sum_{k=1}^{K} -\log p({\bf z}_{t}^{[k]} | {\bf x}_{t}, {\bf m}).
    \label{eq:optimization_sum}
\end{align}
It should be noted that $p({\bf z}_{t}^{[k]} | {\bf x}_{t}, {\bf m})$ is divided by its maximum value for normalization because $p({\bf z}_{t}^{[i]} | {\bf x}_{t}, {\bf m})$ is a probabilistic density function and sometimes its return values exceed 1.

To find the pose satisfying Eq.~(\ref{eq:optimization_sum}), we use the Gauss-Newton method because we would like to efficiently obtain the Hessian matrix, $H$, to approximate the measurement model distribution.
Hence, we re-write Eq.~(\ref{eq:optimization_sum}) using summation of squared residual as
\begin{align}
    \begin{gathered}
        {\bf x}_{t} = \argmin_{{\bf x}_{t}} \frac{1}{2} \sum_{k=1}^{K} \left( 1 - p({\bf z}_{t}^{[k]} | {\bf x}_{t}, {\bf m}) \right)^{2}, \\
        1 - p({\bf z}_{t}^{[k]} | {\bf x}_{t}, {\bf m}) \coloneqq e({\bf x}_{t}, {\bf z}_{t}^{[k]}, {\bf m}).
    \label{eq:optimization_sum_prob_error}
    \end{gathered}
\end{align}
To use the Gauss-Newton method, Jacobian, $J$, is needed to be determined and we need to differentiate the measurement model.
However, the measurement model usually contains undifferentiable factors.
Hence, we use numerical differentiation.
$ij$ element of $J$ is denoted as
\begin{align}
    \begin{gathered}
        J_{ij} = \frac{\partial e({\bf x}_{t}, {\bf z}_{t}^{[i]}, {\bf m})}{\partial x_{j}}
        \simeq \frac{e({\bf x}_{t}, {\bf z}_{t}^{[i]}, {\bf m}) - e({\bf x}_{t}^{[j]}, {\bf z}_{t}^{[i]}, {\bf m})}{\Delta_{j}}, \\
        {\bf x}_{t}^{[j]} = {\bf x}_{t} + \Delta_{j} {\bf o}_{j},
    \end{gathered}
    \label{eq:numerical_differentiation}
\end{align}
where $\Delta_{j}$ is displacement to calculate numerical partial differential regarding $j$th element of ${\bf x}$ and ${\bf o}_{j}$ is an one-hot vector that $j$th element is 1 and size is same to that of ${\bf x}$.
The pose is updated as shown in Eq.~(\ref{eq:gauss_newton_update}) until converged
\begin{align}
    {}^{l+1}{\bf x}_{t} \leftarrow {}^{l}{\bf x}_{t} - \left( {}^{l}J^{\top} {}^{l}J \right)^{-1} {}^{l}J^{\top} {}^{l}{\bf e},
    \label{eq:gauss_newton_update}
\end{align}
where ${\bf e}$ is a residual vector, ${\bf e} = (e^{[1]}, ..., e^{[K]})^{\top}$.
We assume that optimization is converged if $\sum_{k=1}^{K} |{}^{l+1}e^{[k]} - {}^{l}e^{[k]}| / K< \delta$, where $\delta$ is an arbitrary positive constant.

\subsection{Fusion of scan matching via importance sampling}
\label{subsec:fusion_of_sm_and_pf}

Through the optimization shown in Eq.~(\ref{eq:optimization_sum_prob_error}), the pose minimizing the cost function can be obtained.
We refer this pose to an optimized pose, ${\bf x}_{t}^{\rm opt}$.
In addition, we assume that $\left( J^{\top} J \right)^{-1}$ approximates the Hessian matrix, $H$, against the optimized pose and can be used to represent uncertainty of the optimized pose~\cite{BengtssonIROS2001}.
We use this inverse matrix as a covariance matrix and assume that the measurement model can be approximated by the normal distribution
\begin{align}
    p({\bf z}_{t} | {\bf x}_{t}, {\bf m}) \simeq \mathcal{N} \left( {\bf x}_{t}; {\bf x}_{t}^{\rm opt}, \frac{1}{{}^{\rm O}\sigma^{2}}H^{-1} \right),
    \label{eq:approximated_measurement_model}
\end{align}
where ${}^{\rm O}\sigma^{2}$ is a scale factor and must be adjusted based on experiments.

From the normal distribution, pose samples can be drawn.
Hence, Eq.~(\ref{eq:approximated_measurement_model}) can be used as a proposal distribution for particles.
The sampled poses from Eq.~(\ref{eq:approximated_measurement_model}) can be fused with PF used for calcilating Eq.~(\ref{eq:recursive_bayes_filter}) via importance sampling.
Likelihood of the sampled particles, ${}^{\rm O}\omega_{t}^{[i]}$, is determined as
\begin{align}
    \begin{gathered}
        {}^{\rm O}\omega_{t}^{[i]} = \int \int p({}^{\rm O}{\bf x}_{t}^{[i]} | {\bf x}_{t-1}, {\bf x}_{t-2}) p({\bf x}_{t-1} | {\bf x}_{t-2}, {\bf z}_{1:t-1}, {\bf m}) \\
        \cdot p({\bf x}_{t-2} | {\bf z}_{1:t-2}, {\bf m}) {\rm d}{\bf x}_{t-2} {\rm d}{\bf x}_{t-1},
        \label{eq:optimized_particle_likelihood}
    \end{gathered}
\end{align}
where ${}^{\rm O}{\bf x}_{t}^{[i]}$ is a sampled pose.
Eq~(\ref{eq:optimized_particle_likelihood}) means that likelihood is determined according to the particle distribution updated by the linear interpolation.
To calculate these likelihoods, we approximate the distributions with the particles and determine numerical continuous distribution.

In the presented PF, there are two particle distributions sampled from the predictive distribution and measurement model and these likelihoods are calculated using the measurement model and predictive distribution, respectively.
These particle distributions individually approximates the posterior, $p({\bf x}_{t} | {\bf z}_{1:t}, {\bf m})$, and relative difference of likelihood cannot be known if the particles sampled by the predictive distribution are evaluated using the decomposed measurement model, $\prod_{k=1}^{K} p({\bf z}_{t}^{[k]} | {\bf x}_{t}, {\bf m})$.
This yields a problem that the particles cannot be appropriately re-sampled from the both distributions.
To avoid this problem, likelihood of the particles sampled by the predictive distribution is calculated using the approximated measurement model shown in Eq.~(\ref{eq:approximated_measurement_model}).

\section{Implementation}
\label{sec:implementation}

This section details how the presented method is implemented.
The presented method is composed of following processes.
\begin{description}
    \item[A] Initialization
    \item[B] Prediction of the particle poses
    \item[C] Measurement model optimization
    \item[D] Sampling from the measurement model
    \item[E] Likelihood calculation for two particle distributions
    \item[F] Pose estimation
    \item[G] Re-sampling if necessary
    \item[H] Go back to B
\end{description}
Following subsections detail each process.

\subsection{Initialization}

Each particle contains a pose, ${\bf x}$, and likelihood, $\omega$.
The pose is composed of a 3D position and Euler angles.
We assume that an initial pose is given and the particles are randomly distributed around the given pose for initialization.
Likelihood of all the particles are uniformly set, i.e., likelihood is set to $1 / M$ if number of particles is $M$.

\subsection{Prediction of the particle poses}

If the linear interpolation is used to predict the particle poses, the particle poses are updated as
\begin{align}
    \hat{{\bf x}}_{t}^{[i]} = {\bf x}_{t-1}^{[i]} + \Delta {\bf x}_{t}^{[i]}, ~
    \Delta {\bf x}_{t}^{[i]} \sim \mathcal{N} \left( {\bf x}_{t-1} - {\bf x}_{t-2}, {}^{\rm M}\Sigma \right),
    \label{eq:pose_prediction}
\end{align}
where ${}^{\rm M}\Sigma$ is a matrix that represents moving uncertainty and all the elements must be positive.
If INS is available, related elements are updated based on a robot's motion model.
In the open software, the differential drive and omni directional models are supported and IMU can be used, but we do not use INS for the presented method in this paper.

\subsection{Measurement model optimization}
\label{sec:measurement_model}

As described in Section~\ref{subsec:formulation}, the class conditional measurement model presented in our previous work~\cite{AkaiIROS2018} is used.
The measurement model is also decomposed as shown in Eq.~(\ref{eq:optimization_prod}) and we need to determine how $p({\bf z}_{t}^{[k]} | {\bf x}_{t}, c_{t}^{[k]}, {\bf m}) p(c_{t}^{[k]})$ is modeled with ${\rm known}$ and ${\rm unknown}$ conditions.

The model with ${\rm known}$ condition is modeled as
\begin{align}
    p({\bf z}_{t}^{[k]} | {\bf x}_{t}, c_{t}^{[k]} = {\rm known}, {\bf m}) = \mathcal{N}(d_{t}^{[k]}; 0, \sigma^{2}),
    \label{eq:known_measurement_model}
\end{align}
where $d_{t}^{[k]}$ is the minimum distance from $k$th measurement point to mapped obstacles and is obtained from the distance field, and $\sigma^{2}$ is a variance.
To efficiently obtain $d_{t}^{[k]}$, the distance field representation method presented in~\cite{AkaiIV2020} is used.
It should be noted that the normal distribution containing a value obtained from the distance field is not differentiable and we use numerical differentiation for the optimization as shown in Eq.~(\ref{eq:numerical_differentiation}).

The model with ${\rm unknown}$ condition is modeled using the exponential distribution
\begin{align}
    p({\bf z}_{t}^{[k]} | {\bf x}_{t}, c_{t}^{[k]} = {\rm unknown}, {\bf m}) = \frac{\lambda \exp(-\lambda r_{t}^{[k]})}{1 - \exp(-\lambda r_{\rm max})},
\end{align}
where $\lambda$ is the hyperparameter and $r_{\rm max}$ and $r_{t}^{[k]}$ are the maximum and $k$th measurement ranges.
The prior $p(c_{t}^{[k]})$ is determined as uniform, i.e., $p(c_{t}^{[k]} = {\rm known}) = p(c_{t}^{[k]} = {\rm unknown}) = 0.5$, since there are no information to infer it.

Finally, the optimization process determines the pose as shown in Eq.~(\ref{eq:optimization_implementation})
\begin{align}
    {\bf x}_{t} = \argmin_{{\bf x}_{t}} \frac{1}{2} \sum_{k=1}^{K} \left( 1 - \sum_{c_{t}^{[k]} \in \mathcal{C}} p({\bf z}_{t}^{[k]} | {\bf x}_{t}, c_{t}^{[k]}, {\bf m}) p(c_{t}^{[k]}) \right)^{2}.
    \label{eq:optimization_implementation}
\end{align}
Note that if the value inside the brackets of Eq.~(\ref{eq:optimization_implementation}) exceeds a threshold, $\epsilon$, that value is ignored in the optimization.
In addition, the voxel grid filter with resolution $r$ is applied to 3D LiDAR measurements before performing optimization to reduce the computational cost.

\subsection{Sampling from the measurement model}

Through the measurement model optimization, we can obtain Jacobian against the optimized pose, ${\bf x}_{t}^{{\rm opt}}$.
The Hessian matrix is calculated using Jacobian and we approximate the measurement model by the normal distribution as shown in Eq.~(\ref{eq:approximated_measurement_model}).
The Cholesky decomposition is applied to $\frac{1}{\sigma^{2}}H^{-1}$ and it is decomposed to $PP^{\top}$.
Pose samples are generated from the measurement model as
\begin{align}
    {}^{\rm O}{\bf x}_{t}^{[i]} = P {\bf t}^{[i]} + {\bf x}_{t}^{{\rm opt}}, ~~~ {\bf t}^{[i]} \sim \mathcal{N}(0, I),
    \label{eq:opt_particle_pose}
\end{align}
where $I$ is identity.
Number of $L$ particles are sampled from the measurement model.

\subsection{Likelihood calculation}

As mentioned in Section~\ref{subsec:fusion_of_sm_and_pf}, there are two particle distributions.
Likelihood of the particles sampled from the predictive distribution, i.e., shown in Eq.~(\ref{eq:pose_prediction}), is calculated as
\begin{align}
    \omega_{t}^{[i]} = \mathcal{N}({\bf \hat{x}}_{t}^{[i]}; {\bf x}_{t}^{{\rm opt}}, \frac{1}{{}^{\rm O}\sigma^{2}}H^{-1}).
    \label{eq:particle_likelihood}
\end{align}
Likelihood of the particles sampled from the measurement model, i.e., shown in Eq.~(\ref{eq:opt_particle_pose}), is calculated as
\begin{align}
    {}^{\rm O}\omega_{t}^{[i]} = \frac{1}{M} \sum_{j=1}^{M} \mathcal{N}({}^{\rm O}{\bf x}_{t}^{[i]}; {\bf \hat{x}}_{t}^{[j]}, {}^{\rm P}\Sigma),
    \label{eq:opt_particle_likelihood}
\end{align}
where ${}^{\rm P}\Sigma$ is an arbitrary covariance matrix.
Eq.~(\ref{eq:opt_particle_likelihood}) means that the predictive distribution is approximated by the Gaussian mixture model.

\subsection{Pose estimation}

Likelihoods are normalized before the pose estimation
\begin{align}
    \begin{gathered}
        \omega_{t}^{{\rm sum}} = \sum_{i=1}^{M} \omega_{t}^{[i]} + \sum_{i=1}^{L} {}^{\rm O}\omega_{t}^{[i]}, \\
        \omega_{t}^{[i]} \leftarrow \frac{\omega_{t}^{[i]}}{\omega_{t}^{{\rm sum}}}, ~~~ {}^{\rm O}\omega_{t}^{[i]} \leftarrow \frac{{}^{\rm O}\omega_{t}^{[i]}}{\omega_{t}^{{\rm sum}}}.
    \end{gathered}
    \label{eq:normalization}
\end{align}
Then, a weighted average pose is calculated and is used as an estimated pose
\begin{align}
    {\bf x}_{t} = \sum_{i=1}^{M} \omega_{t}^{[i]} {\bf x}_{t}^{[i]} + \sum_{i=1}^{L} {}^{\rm O}\omega_{t}^{[i]} {}^{\rm O}{\bf x}_{t}^{[i]}.
\end{align}

\subsection{Re-sampling}

Effective sample size shown in Eq.~(\ref{eq:ess}) is calculated and re-sampling is performed if its value is less than $(M + L) / 2$.
\begin{align}
    \frac{1}{\sum_{i=1}^{M} (\omega_{t}^{[i]})^{2} + \sum_{i=1}^{L} ({}^{\rm O}\omega_{t}^{[i]})^{2}}
    \label{eq:ess}
\end{align}
In the re-sampling process, number of $M$ particles are selected from two particle distributions that approximate the predictive distribution and measurement model according to their likelihood.

\section{Experiments}
\label{sec:experiments}

\subsection{Dataset-based experiments}

We use the SemanticKITTI dataset~\cite{behley2019iccv} to validate the presented method.
The SemanticKITTI dataset contains 3D LiDAR points with semantic labels such as car and building.
The 3D LiDAR points obtained from static objects, e.g., building and road, are mapped according to the ground truth trajectory and the mapped points are used as a localization map.
In the localization phase, all 3D LiDAR points without semantic labels are feed to the localization module and it performs ego-vehicle pose tracking.
The estimated poses are compared with the ground truth and position and orientation errors, $\sqrt{\Delta x^{2} + \Delta y^{2} + \Delta z^{2}}$ and $\sqrt{\Delta \varphi^{2} + \Delta \theta^{2} + \Delta \psi^{2}}$, are calculated, where $x$, $y$, and $z$ are 3D position and $\varphi$, $\theta$, and $\psi$ are roll, pitch, and yaw angles.

The localization methods using the measurement model optimization do not use INS, but a method described in Section~\ref{subsubsec:spf} uses INS for ensuring localization performance.
Moving amount is calculated from the consecutive ground truth poses.
Noises are added to the moving amount and odometry is simulated.
This simulated odometry is used as INS.

\subsection{Comparison methods in dataset-based experiments}

\subsubsection{Measurement model optimization (MMO)}

This method estimates a pose satisfying Eq.~(\ref{eq:optimization_implementation}).
The initial pose for the optimization is estimated using linear interpolation from two previous estimates.

To validate the class conditional measurement model presented in~\cite{AkaiIROS2018}, we also use MMO with the likelihood field model~\cite{Thrun:2005:PR:1121596}.
This method is referred to MMOLFM.

\subsubsection{Standard particle filter (SPF)}
\label{subsubsec:spf}

In SPF, particle poses are predicted using the predictive distribution, i.e., particle poses are updated using odometry.
$z$, $\varphi$, and $\psi$ are predicted using the linear interpolation.
Then, particle likelihoods are calculated with the measurement model shown in Eq.~(\ref{eq:class_conditional_measurement_model}).

\subsubsection{Extended Kalman filter (EKF)}

The pose is predicted using the linear interpolation and pose covariance is also estimated.
Then, MMO is performed.
The covariance regarding MMO estimate is determined as shown in Eq.~(\ref{eq:approximated_measurement_model}).
The predicted and estimated poses are fused with Kalman filter.

\subsection{Parameters}

Table~\ref{tab:parameters} summarizes the parameters used in the dataset-based experiments.
The same parameters used in all the methods.
Note that $z_{{\rm hit}}$, $z_{{\rm rand}}$, and $z_{{\rm max}}$ are the parameters used in the likelihood field model~\cite{Thrun:2005:PR:1121596}.

\begin{table}[!t]
\begin{center}
\caption{Parameters used in the experiments}
\begin{tabular}{|cc|c|}
\hline
\multicolumn{1}{|c|}{$M = 1000$}                                           & $L = 1000$                                               & $r = 1$                    \\ \hline
\multicolumn{1}{|c|}{$\left( {}^{\rm M}\Sigma \right)_{ij} = 0.5~(i = j)$} & $\left( {}^{\rm M}\Sigma \right)_{ij} = 0.01~(i \neq j)$ & $\sigma^{2} = 0.4$         \\ \hline
\multicolumn{1}{|c|}{$z_{{\rm hit}} = 0.9$}                                & {$z_{{\rm rand}} = z_{{\rm max}} = 0.05$}                & ${}^{\rm O}\sigma^{2} = 1$ \\ \hline
\multicolumn{1}{|c|}{$\lambda = 0.001$}                                    & $r_{{\rm max}} = 120$                                    & {$ \delta = 0.02$}         \\ \hline
\multicolumn{2}{|c|}{${}^{\rm P}\Sigma = {\rm diag}(0.3, 0.3, 0.3, 0.1, 0.1, 0.1)$}                                                   & $\epsilon = 0.5$           \\ \hline
\end{tabular}
\label{tab:parameters}
\end{center}
\end{table}

\subsection{Results}
\label{subsec:results}

Table~\ref{tab:positional_errors} and \ref{tab:angular_errors} show the positional and angular error averages and their standard deviations, respectively.
Cells without values mean that the localization method could not successfully track the vehicle pose, i.e., the average positional error exceeded 1~m.
The minimum positional and angular average errors are highlighted in bold in each sequence.
The presented localization method is referred to PFF that stands for particle-filter-based fusion.

\begin{table*}[!t]
\begin{center}
\caption{Positional errors in centimeters (average / standard deviation).}
\begin{tabular}{c|c|c|c|c|c|c|c|c} \hline
Sequence & 01 & 03 & 04 & 05 & 06 & 07 & 09 & 10 \\ \hline
MMO & {\bf 29.63 / 33.68} & {\bf 14.74 / 7.26} & 13.48 / 5.98 & 20.56 / 7.78 & 17.78 / 6.12 & 31.66 / 11.71 & {\bf 21.71 / 10.04} & {\bf 19.16 / 8.32} \\ \hline
MMOLFM & - & 21.87 / 9.41 & 15.60 / 6.75 & 25.45 / 12.70 & 20.48 / 8.15 & 34.33 / 13.41 & 27.29 / 14.67 & 27.03 / 11.58 \\ \hline
SPF & - & - & - & 26.63 / 20.57 & - & 43.89 / 30.43 & - & 41.26 / 63.89 \\ \hline
EKF & 30.79 / 38.12 & 14.91 / 7.56 & {\bf 12.81 / 5.69} & {\bf 20.42 / 7.47} & {\bf 17.55 / 6.16} & {\bf 31.35 / 12.05} & 21.73 / 9.64 & 19.40 / 8.74 \\ \hline
PFF & 30.40 / 30.05 & 23.12 / 10.76 & 18.92 / 10.60 & 25.70 / 10.66 & 25.23 / 9.69 & 35.05 / 11.97 & 26.94 / 11.81 & 24.90 / 9.64 \\ \hline
\end{tabular}
\label{tab:positional_errors}
\end{center}
\end{table*}

\begin{table*}[!t]
\begin{center}
\caption{Angular errors in degrees (average / standard deviation).}
\begin{tabular}{c|c|c|c|c|c|c|c|c} \hline
Sequence & 01 & 03 & 04 & 05 & 06 & 07 & 09 & 10 \\ \hline
MMO & 0.76 / 0.25 & 0.58 / 0.20 & 0.68 / 0.15 & {\bf 0.61 / 0.27} & 0.66 / 0.21 & {\bf 0.70 / 0.29} & 0.61 / 0.22 & {\bf 0.75 / 0.45} \\ \hline
MMOLFM & - & 0.86 / 0.29 & 0.88 / 0.20 & 1.11 / 0.66 & 1.12 / 0.36 & 1.09 / 0.41 & 1.00 / 0.38 & 1.13 / 0.61 \\ \hline
SPF & - & - & - & 1.24 / 1.10 & - & 1.97 / 1.93 & - & 1.61 / 1.35 \\ \hline
EKF & 0.76 / 0.25 & 0.58 / 0.19 & 0.65 / 0.19 & 0.62 / 0.23 & 0.69 / 0.21 & 0.71 / 0.30 & 0.61 / 0.22 & 0.77 / 0.45 \\ \hline
PFF & {\bf 0.71 / 0.28} & {\bf 0.56 / 0.21} & {\bf 0.64 / 0.22} & 0.62 / 0.26 & {\bf 0.64 / 0.24} & {\bf 0.70 / 0.33} & {\bf 0.60 / 0.26} & 0.81 / 0.56 \\ \hline
\end{tabular}
\label{tab:angular_errors}
\end{center}
\end{table*}

MMO and EKF accurately worked in all the sequences and PFF could not outperform them in terms of the positional accuracy.
Optimization-based methods are typically accurate more than probabilistic approaches and the results also showed that.
Since odometry-less EKF tends to similar to results by scan matching, it also accurately worked.
The average computational times for an estimation by MMO, MMOLFM, SPF, EKF, and PFF are 32.33, 18.50, 38.81, 31.15, and 48.73~msec, respectively (the used CPU is Intel(R) Core(TM) i9-9820X CPU@3.30~GHz and all the methods were executed on a single CPU thread).
Since the measurement cycle of the 3D LiDAR used in the KITTI dataset is 10~Hz, all the methods worked in real time.
However, PFF could also not outperform MMO and EKF in terms of the computational cost.
This is obvious because PFF also uses MMO in its estimation process and other additional processes are implemented.
These results revealed that PFF is not a better method if we focus on only accuracy and efficiency.

SPF worked in some sequences; however, it failed localization in 5 sequences and its estimation accuracy was worst even it achieved the vehicle tracking.
To improve performance of SPF, many particles must be used; however, it increases the computational cost and makes difficult to achieve real time localization.
For example, to use 10000 particles in SPF, computational time for an estimation was to approximately 250~msec in our implementation.
In addition, SPF requires INS for accurate estimation.
However, PFF does not require the use of INS and achieved real time 3D localization on a single CPU.
From these results, we consider that the presented method is an efficient solution to 3D-LiDAR-based MCL.
In addition, we consider that the presented method enables to introduce probabilistic methods that improves localization performance such as presented in our previous work~\cite{AkaiJFR2023} in 3D LiDAR localization.

Additionally, we would like to emphasize that MMO outperformed MMOLFM in all the sequences.
This result revealed that the class conditional measurement model contributes to improve localization performance in non static environments.

\subsection{Comparison with other methods}

We further conducted comparison with other two methods for qualitative evaluation.
Because the used logs do not have the ground truth data, this subsection just shows difference of performances.

\subsubsection{Comparison with HDL localization}

\begin{figure}[!t]
    \begin{center}
        \includegraphics[width = 85 mm]{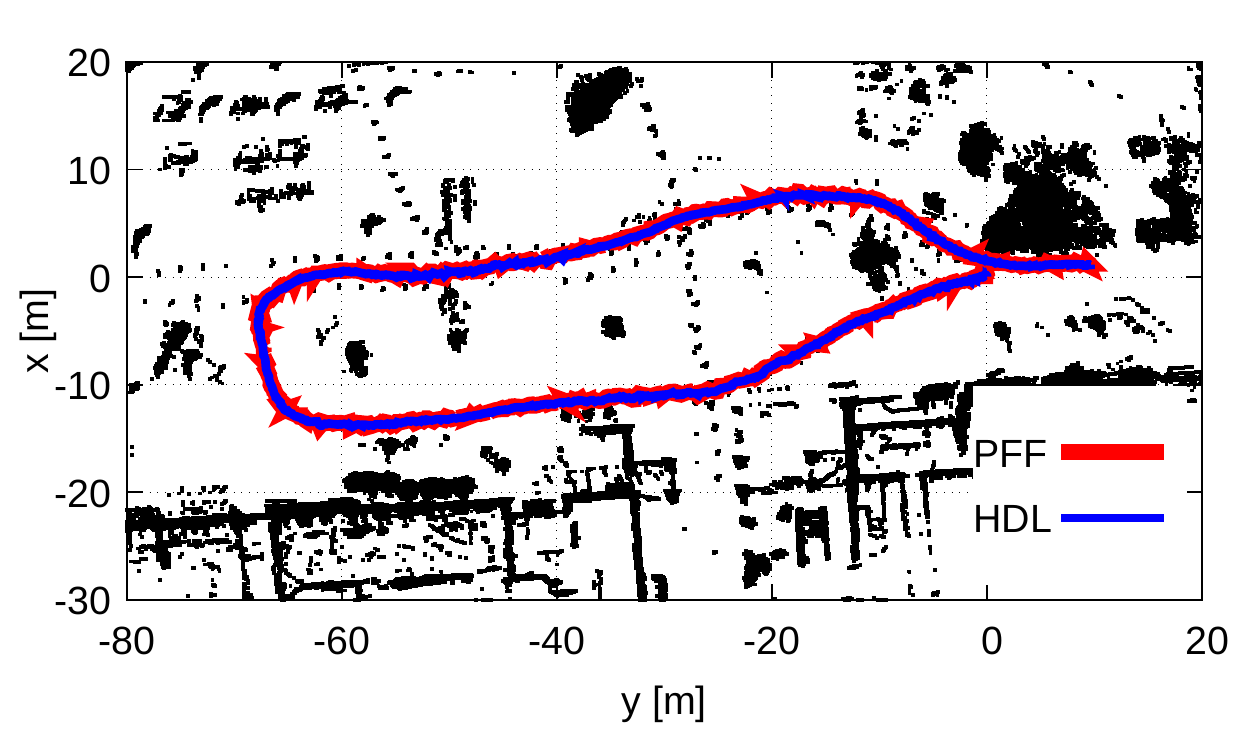}
        \caption{Comparison with HDL localization~\cite{hdl_localization}. The black points depict the point cloud map.}
        \label{fig:hdl_trajectories}
    \end{center}
\end{figure}

HDL localization presented in~\cite{hdl_localization} is 3D-LiDAR-based localization and works without INS.
We used the sample map and log data provided in its GitHub page\footnote{\url{https://github.com/koide3/hdl_localization}} for the comparison.
Figure~\ref{fig:hdl_trajectories} shows a comparison result.
PFF estimate was not smooth more than that of HDL localization.
HDL localization performs unscented-Kalman-filter-based fusion and achieves smooth trajectory estimation.
PFF also performs PF-based fusion, but smooth trajectory could not be estimated owing to random sampling of the particles.
In addition, fusion of two particle distributions via importance sampling loses robustness more than EKF-based fusion in our implementation.
This is a limitation of the presented method.

However, PFF outperformed HDL localization in terms of computational efficiency.
HDL localization performs multi-threaded NDT scan matching.
In our experiment, we used i9-9820X as CPU and it has 20 threads.
Approximate CPU load of HDL localization was 300~\% when downsampling scale for sensor measurement points was set to 1~m and used all the threads.
Instead, approximate CPU load of PFF was 60~\% in the same condition.
It should be noted that HDL localization has an option to reduce the computational cost, but we could not succeed localization if the option was used in our experiment.

\begin{figure}[!t]
    \begin{center}
        \includegraphics[width = 85 mm]{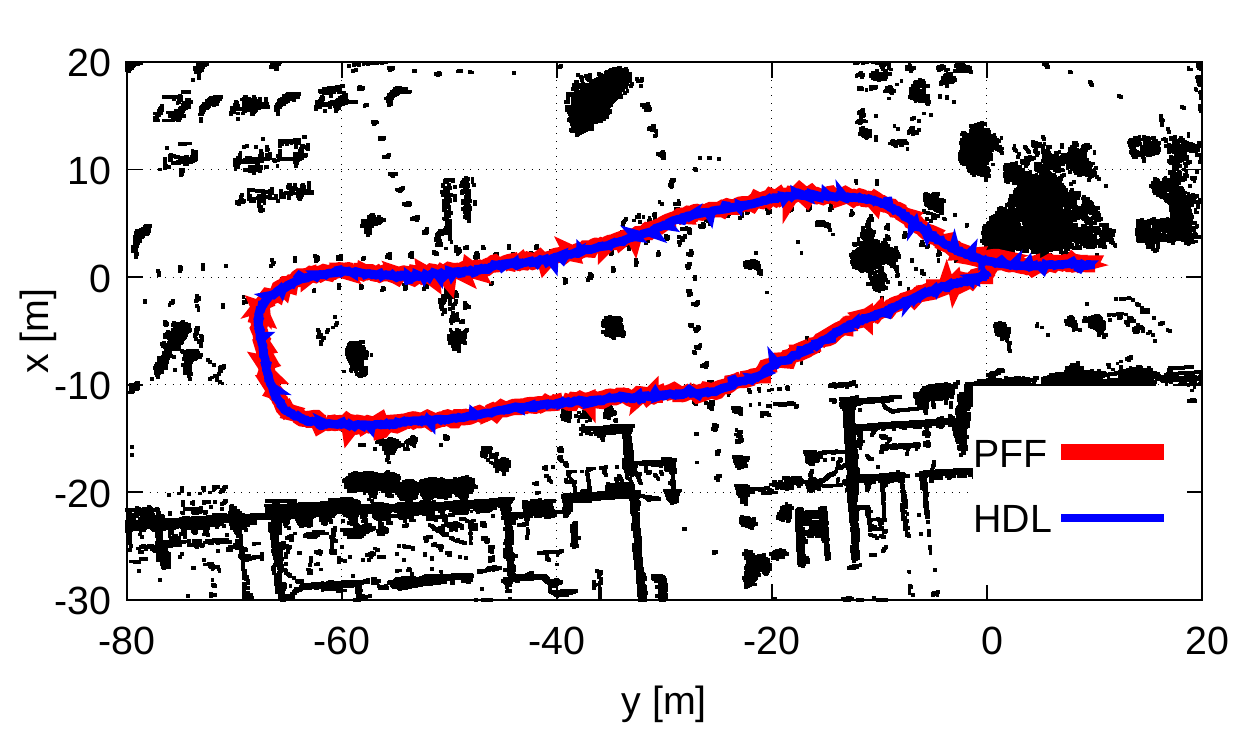}
        \caption{Comparison when the optimization results were disturbed.}
        \label{fig:hdl_trajectories_noise}
    \end{center}
\end{figure}

Figure~\ref{fig:hdl_trajectories_noise} also shows a comparison result with HDL localization.
In this comparison, arbitrary noise was added to the position estimated by the optimization-based method, i.e., the measurement model optimization and multi-threaded NDT SM, every one second.
Even when the optimization results were disturbed, both PFF and HDL localization achieved robust pose tracking.
We confirmed that PFF has similar performance to unscented-Kalman-filter-based fusion.

\subsubsection{Comparison with mcl\_3dl}

\begin{figure}[!t]
    \begin{center}
        \includegraphics[width = 85 mm]{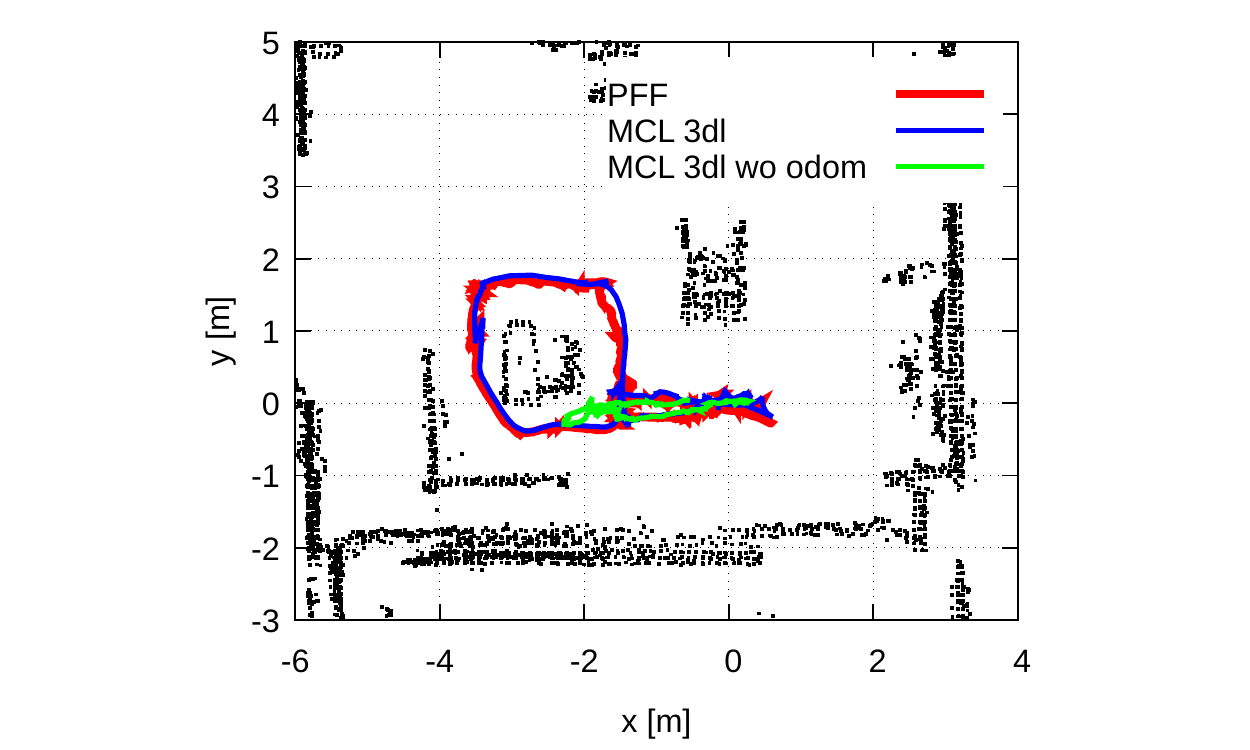}
        \caption{Comparison with mcl\_3dl.}
        \label{fig:mcl_3dl_trajectories}
    \end{center}
\end{figure}

We also compared with mcl\_3dl that is MCL for 3D localization and is an open source software\footnote{\url{https://github.com/at-wat/mcl_3dl}}.
This package also provides the sample map and log data and we used them for the comparison.
Figure~\ref{fig:mcl_3dl_trajectories} shows a comparison result.
mcl\_3dl requires odometry and/or IMU and we tested two cases where both odometry and IMU are used (MCL 3dl) and IMU is only used (MCL 3dl wo odom).
mcl\_3dl successfully worked when odometry was used, but it failed localization if odometry is not available.
However, PFF achieved localization even if INS is not used.
mcl\_3dl is similar to SPF described in Section~\ref{subsubsec:spf} and similar results shown in Section~\ref{subsec:results} that is SPF needs INS for accurate localization were obtained.
However, smoothness of the estimated trajectory by PFF was also not smooth more than that of mcl\_3dl owing to less of INS.
It should be noted that the parameters depending on the maximum measurement range shown in Table~\ref{tab:parameters} were only modified as $\lambda = 0.01$ and $r_{{\rm max}} = 30, r = 0.2$.

\subsection{Discussion}
\label{subsec:discussion}

The main contribution of this work is presenting the method to efficiently perform 3D-LiDAR-based MCL.
SPF of course can be applied to 3D MCL; however it requires large computational cost.
The presented method does not require such large cost and achieves accurate localization.
Consequently, the simultaneous estimation methods that we presented in~\cite{AkaiJFR2023} can be implemented in the 3D localization case by using the presented method.
Showing effectiveness of the methods such as reliability estimation, unknown obstacle estimation, and failure recovery in the 3D localization case is our future work and we believe that it contributes realize a more reliable 3D localization method.
However, we also would like to discuss weak points of the presented method in this subsection.

First, the presented solution cannot have high non-linearity more than SPF.
In SPF, likelihood of the particles can be calculated using any types of model and can have high non-linearity.
However, likelihood of the particles used in the presented method is calculated using the approximated measurement model and predictive distribution as shown in Eqs.~(\ref{eq:particle_likelihood}) and (\ref{eq:opt_particle_likelihood}).
As a result, high non-linearity is lost and it decreases robustness of PF.
Hence, PFF cannot outperform SPF if significant accurate INS and/or a high power computer is available.
However, the use of such devices is not practical for real world application.

Ensuring high-nonlinearly could be possible even if PFF is used.
However, it requires effective approximation for the measurement model.
This approximation must need to achieve effective sampling of poses and likelihood calculation.
Achieving such an approximation is also a difficult task and we could not realize that currently.
It is known that a neural network (NN) with Monte Carlo dropout can approximate posterior over an output conditioned on an input~\cite{Gal2015BayesianCN} and we used it to sample particles from the measurement model in~\cite{AkaiICRA2020}.
We consider that this technique could be utilized to overcome the problem; however, we also need to consider how accurately to approximate the measurement model using NN.
For example, the measurement model has a map as a condition and this means that a map is needed to be fed to NN and this makes a concern that how to handle large map data.
In addition, we need to care robustness and stability of NN as presented in~\cite{sattler19cvpr}.

The presented method could not achieve smooth trajectory estimation.
The method used two particle distributions and we performed re-sampling from the both distributions.
If EKF-based fusion is used, a noisy estimate can be ignored in the fusion process; however, it is difficult for the presented method to have similar effect.
To appropriately re-sample the particles from the both distribution, we calculated likelihood of the particles drawn from the predictive distribution using the approximated measurement model.
Consequently, the particles drawn from the predictive distribution cannot have likelihood that explicitly considers sensor measurements and might not have suitable likelihood in some cases.
This is also a problem owing to loss of non-linearity.

\section{Conclusion}
\label{sec:conclusion}

This paper has presented an efficient solution to 3D-LiDAR-based MCL.
The presented method has two importance processes; optimization of the measurement model and fusion of it with MCL via importance sampling.
The measurement model optimization is performed using numerical differentiation because the measurement model contains non differentiable factors.
To efficiently perform the optimization, we used the efficient distance field representation method presented in our previous work~\cite{AkaiIV2020}.
Owing to the measurement model optimization, any types of measurement models can be approximated by the normal distribution and this distribution can be used to sample the particles, i.e., a new proposal distribution.
Consequently, particles can be exactly drawn around the ground truth even if INS is not available.
In addition, this approximation enables us to easily fuse the result with MCL via importance sampling.
The experimental results showed 3D-LiDAR-based localization can be solved using PF without any heuristic techniques except for some approximation.
As far as we know, this is the first study that realizes 3D-LiDAR-based MCL without INS and any acceleration techniques such as GPU.
In addition, the presented method obeys on rigorous probabilistic manners.
We would like to extend this solution to other simultaneous estimation methods such as we presented previously~\cite{AkaiJFR2023} in the future.





%

\section*{ACKNOWLEDGMENT}

This work was supported by KAKENHI under Grant 40786092.

\bibliographystyle{unsrt}
\bibliography{root.bib}


\end{document}